\documentclass[conference]{IEEEtran}
\usepackage{cite}
\usepackage{amsmath,amssymb,amsfonts}
\usepackage{algorithmic}
\usepackage{graphicx}
\usepackage{textcomp}
\usepackage{xcolor}
\usepackage{multirow}
\def\BibTeX{{\rm B\kern-.05em{\sc i\kern-.025em b}\kern-.08em
    T\kern-.1667em\lower.7ex\hbox{E}\kern-.125emX}}
\begin{document}

\title{Benchmarking LLMs for Predictive Applications in the Intensive Care Units\\}

\author{\IEEEauthorblockN{\hspace{-5em}Chehak Malhotra\textsuperscript{\dag}}
\IEEEauthorblockA{\textit{\hspace{-5.5em}Computer Science} \\
\textit{\hspace{-5em}Indraprastha Institute of Information Technology Delhi}\\
\hspace{-5em}Delhi, India \\
\hspace{-5em}chehak21141@iiitd.ac.in}
\and
\IEEEauthorblockN{Mehak Gopal \textsuperscript{\dag}}
\IEEEauthorblockA{\textit{Computational Biology} \\
\textit{Indraprastha Institute of Information Technology Delhi}\\
Delhi, India \\
mehak21475@iiitd.ac.in}
\and
\IEEEauthorblockN{Akshaya Devadiga}
\IEEEauthorblockA{\textit{Computational Biology} \\
\textit{Indraprastha Institute of Information Technology Delhi}\\
Delhi, India \\
0000-0002-2481-6528}
\and
\IEEEauthorblockN{Pradeep Singh}
\IEEEauthorblockA{\textit{Computational Biology} \\
\textit{Indraprastha Institute of Information Technology Delhi}\\
Delhi, India \\
0000-0002-1215-7553}
\and
\IEEEauthorblockN{Ridam Pal}
\IEEEauthorblockA{\textit{Computational Biology} \\
\textit{Indraprastha Institute of Information Technology Delhi}\\
Delhi, India \\
ridamp@iiitd.ac.in}
\and
\IEEEauthorblockN{Ritwik Kashyap}
\IEEEauthorblockA{\textit{Computational Biology} \\
\textit{Indraprastha Institute of Information Technology Delhi}\\
Delhi, India \\
ritwik21485@iiitd.ac.in}
\and
\IEEEauthorblockN{Dr. Tavpritesh Sethi\textsuperscript{*}}
\IEEEauthorblockA{\textit{Computational Biology} \\
\textit{Indraprastha Institute of Information Technology Delhi}\\
Delhi, India \\
tavpriteshsethi@iiitd.ac.in}

\thanks{\textsuperscript{\dag}Equal contribution}
\thanks{\textsuperscript{*}Corresponding author}
}

\maketitle

\begin{abstract}
With the advent of LLMs, various tasks across the natural language processing domain have been transformed. However, their application in predictive tasks remains less researched. This study compares large language models, including GatorTron-Base (trained on clinical data), Llama 8B, and Mistral 7B, against models like BioBERT, DocBERT, BioClinicalBERT, Word2Vec, and Doc2Vec, setting benchmarks for predicting Shock in critically ill patients. Timely prediction of shock can enable early interventions, thus improving patient outcomes.  Text data from 17,294 ICU stays of patients in the MIMIC III database were scored for length of stay $>$ 24 hours and shock index (SI) $>$ 0.7 to yield 355 and 87 patients with normal and abnormal SI-index, respectively. Both focal and cross-entropy losses were used during fine-tuning to address class imbalances.
Our findings indicate that while GatorTron Base achieved the highest weighted recall of 80.5\%, the overall performance metrics were comparable between SLMs and LLMs. This suggests that LLMs are not inherently superior to SLMs in predicting future clinical events despite their strong performance on text-based tasks. To achieve meaningful clinical outcomes, future efforts in training LLMs should prioritize developing models capable of predicting clinical trajectories rather than focusing on simpler tasks such as named entity recognition or phenotyping.

\end{abstract}

\begin{IEEEkeywords}
electronic health record, intensive care unit, large language models, small language models, machine learning, shock index
\end{IEEEkeywords}

\section{Introduction}
Since the introduction of Large Language Models (LLMs) in 2018 \cite{bert}, researchers have been exploring their capabilities across varied domains. Recent breakthroughs with human-level conversations with GPT models have led to enthusiastic applications of LLMs for clinical tasks \cite{review} \cite{review2}. However, it remains unclear whether LLMs have a better understanding of the clinical text than Small Language Models (SLMs), particularly in tasks involving the modeling of patient trajectories, which is a relatively unexplored area.  In a clinical setting, designing early warning systems for future complications is extremely useful for practitioners to help prevent adverse events, mitigate risks, and implement timely interventions to improve patient outcomes and reduce mortality rates \cite{predllm}.  This is particularly invaluable in Intensive Care Units (ICUs), which handle severe life-threatening cases. 

This study aims to evaluate the potential of LLMs in predictive applications in critical care settings. For our study, we focus on evaluating LLMs to predict the onset of shock. Shock is a manifestation of physiological decompensation, and the shock index serves as a tool to quantify the degree of this decompensation. Shock is a critical condition characterized by insufficient tissue perfusion and oxygenation, leading to organ dysfunction and high mortality rates if untreated
\cite{hschok}\cite{icu}. Despite advancements in critical care, predicting physiological decompensation remains a complex challenge due to patient data's dynamic nature in ICUs. 

The increasing availability of electronic health records (EHRs) has fueled the development of predictive models to augment decision-making in ICUs. These models leverage structured data, such as vital signs and laboratory results, and unstructured clinical notes to identify early warning signals. For example, BERT-based models have been successfully employed to predict shock onset by extracting features from physician notes and integrating them with physiological data \cite{bertintro}. These approaches have demonstrated the utility of language models in processing unstructured clinical data to improve predictive accuracy.

Compared with BERT and other transformer-based architectures, LLMs are expected to offer more advanced contextual understanding and broader pretraining across diverse datasets. In this study, we hypothesized LLMs may yield better predictions of physiological decompensation and aimed to (i) assess the feasibility of using LLMs for shock prediction in ICUs and (ii) compare the performance against SLMs.

\section{Method}

\subsection{Data curation and Cohort creation}
The data used for this study was curated using the publicly available Medical Information Mart for Intensive Care (MIMIC) III v.1.4 database \cite{mimic}\cite{mimic2}.  This data provides de-identified health-related data, from which 17,294 ICU stays were used for our analysis. Note events data was used to create 24-hour cohorts for multiple patients using physician notes and vitals. In the critical care environment, the patient's heart rate (HR), respiratory rate (RR), systolic blood pressure (SBP), and oxygen saturation (SpO2) were also retrieved for every 24 hours.

Shock Index (SI) was defined for each patient entry as HR divided by SBP to create the cohort  \cite{shockindex}.  The patient cohort utilized in this study was derived from the cohort construction described in reference \cite{shcokmode}. The continuous vitals data was summarized at 1-minute intervals and utilized to calculate the SI. A median SI was calculated using time epochs of 30 minutes from the vitals. Each epoch was labeled abnormal if the SI $\geq 0.7$, normal if the SI $<$ 0.7, or NA if no data was present. Days without an SI label (NA) were removed. Any episode of abnormal SI lasting longer than thirty minutes that was preceded by at least twenty-four hours of normal SI was considered a new occurrence of abnormal SI. Only new instances of abnormal SI were considered to avoid trivial predictions via ongoing abnormal SI. For the textual data, the top seven frequencies of physician notes type were considered, i.e., notes from physician residents, intensivists, the physician attending, and other ICU notes. Physician notes were mapped using date, subject ID, and ICU stay ID to their next-day labels generated using continuous vitals data. The textual cohort for our experiments had 355 normal SI patients and 87 abnormal SI patients.

\subsection{Data Preprocessing}
The textual data in physician notes was formatted before any other preprocessing step. The encrypted information ([**word**]) and single-letter words were removed. All text was converted to lowercase and underwent standard pre-processing, such as removal of whitespaces, stopwords, punctuations, digits, and words with less than or equal to two letters. The misspelled words were rectified using Levenshtein distance \cite{b5}, and a clinician vetted the resulting vocabulary. Words directly indicative of the onset of abnormal shock index in physician notes, specifically, diagnoses such as ‘shock’ and ‘septic,’ along with therapeutics such as ‘dobutamine,’ ‘dopamine,’ ‘adrenaline,’ and ‘noradrenaline’ were masked.
The therapeutics were extracted using Named Entity Recognition (NER) via MED7 \cite{med7}, the feature space. Therapeutics with more than one word was concatenated using an underscore character. 

\subsection{Feature extraction using LLMs}
Our experiments utilized three LLMs: a clinical LLM, Gatortron Base \cite{gator}, and a general purpose LLM, Llama-3.1-8B-Instruct \cite{llama} and Mistral-7B Instruct v0.3 \cite{mistral}.

A history of present illness (HOPI) was also considered in the clinical context for each patient and the therapeutics. HOPI comprises the chief complaint, assessment and plan, and patient history. This gave us a comprehensive overview of the patient’s current issues, symptom progression, and duration in detail and the provisional management schedule based on the initial clinical evaluation. Based on this, a common feature set was considered for each patient, which included a numerical value (i.e., the embedding) for each therapeutic in the entire data frame and HOPI. 

Each therapeutic was analyzed within a context window of 512 words, centered on the drug itself, with 256 words on either side to capture the surrounding context.  The mean of these embeddings was considered the numerical representation of each therapeutic. 

For HOPI, the length of the text exceeded the maximum input constraint of 512 tokens for Gatortron; hence, to get an adequate representation for these, the text was split into chunks, and the [CLS] tokens were aggregated via mean pooling. 

For Llama and Mistral, the input constraints were extended to 4096 words for both therapeutics and HOPI, leveraging their capability to process longer sequences. This entire feature vector was considered for further model validation.

\subsection{Model Fine Tuning}
To fine-tune Gatortron, we utilized two loss functions: cross-entropy and focal loss. Focal loss \cite{focal} \eqref{eq:focal_loss} is a modified version of cross-entropy loss \cite{ce} and addresses class imbalance. 
\begin{equation}
FL(p_t) = -\alpha_t (1 - p_t)^\gamma \log(p_t)\label{eq:focal_loss}
\end{equation}

For our experiments, we set our focusing parameter, $\gamma$, as 2, focusing our loss on more complex examples.
A combined feature set, including therapeutics and HOPI, was used for each patient. Given the token limitations of the LLM, the input data was limited to the first 512 tokens per instance. A stratified split of 70:20:10 (training, testing, and validation) was considered, where all multiple instances of a patient ID were considered in one dataset.  
A model was fine-tuned for 20 epochs for each loss and considered for further classification analysis. 

For our models using both focal loss and cross-entropy loss, we observed that focal loss consistently resulted in lower overall loss trends than cross-entropy. During the validation phase, we noticed an increasing trend in loss across epochs. To mitigate this, we conducted experiments with dropout regularization at various probabilities (0.3, 0.5, 0.7, 0.9999, 0.999999). Based on the validation losses observed across each epoch, we selected three models that demonstrated the most effective performance with dropout applied to the focal loss framework.

\subsection{Classification models development }
A stratified split of 80:20 was considered for training and testing. All multiple instances of a patient ID were considered in one data set. 
For our fine-tuned models, the validation and test set were considered for the classification models to prevent overlap in data. 
For classification purposes, we used five models: Gradient Boost, Random Forest, XGBoost, Logistic Regression, and AdaBoost. 
Due to the extensive number of features present, feature selection using ExtraTreesClassifier \cite{b1} was performed, and the top $100$ features were chosen. To handle the class imbalances, the Synthetic Minority Over-sampling Technique (SMOTE) \cite{b2} was implemented. Most model parameters were set to default values, with select hyperparameters (AdaBoost- learning rate 0.1, XGBoost - n\_estimators 50) explicitly tuned for our classification task. We calculated the average for all metrics across 100 bootstrap iterations, considering all metrics as weighted, except for accuracy.

\begin{figure}[htbp]
\centerline{\includegraphics[width=1\linewidth]{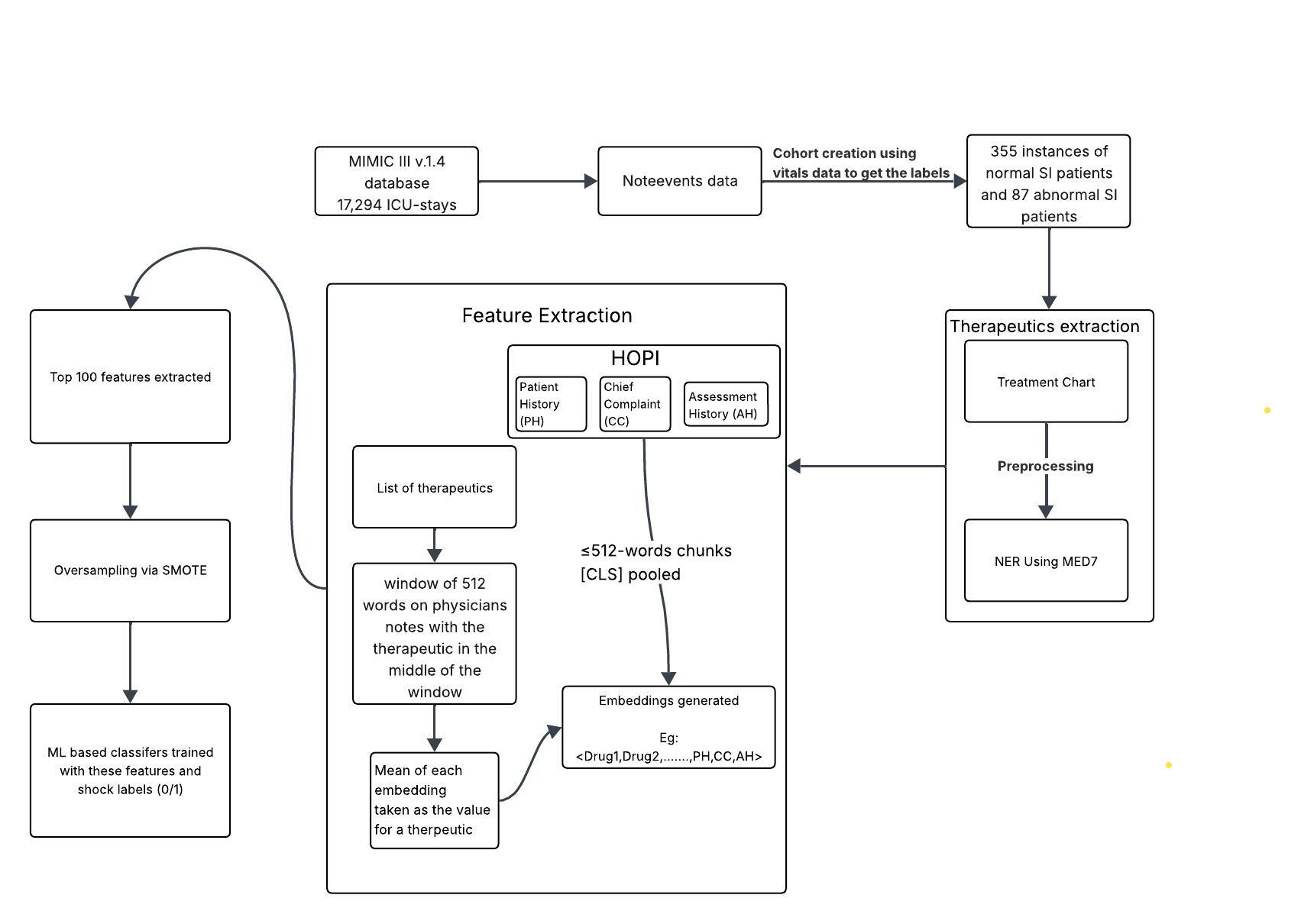}}
\caption{Data pipeline}
\end{figure}

\section{Results}
The following sections provide an overview of various experimental methods undertaken. 

\subsection{Comparative Analysis of LLMs and SLMs}
Table 1 provides a performance comparison of LLMs (GatorTron, Llama, and Mistral) with SLMs (Word2Vec+Doc2Vec, BioClinicalBERT+DocBERT, BioBERT+DocBERT) from ShockModes \cite{thesis} across various evaluation parameters. 

\begin{table*}[htbp]  
    \caption{Benchmark Results Across Different Models and Datasets}
    \centering
    \resizebox{\textwidth}{!}{
        \renewcommand{\arraystretch}{1.5} 
        \setlength{\tabcolsep}{5pt} 
        \begin{tabular}{|c|c|c|c|c|c|c|}
            \hline
            \textbf{Model} & \textbf{Accuracy} & \textbf{Precision} & \textbf{Recall} & \textbf{F1 Score} & \textbf{AUC ROC} & \textbf{Specificity} \\
            \hline
            \multicolumn{7}{|c|}{\textbf{Word2Vec + Doc2Vec}} \\
            \hline
            Logistic Regression & $0.56 \pm 0.0103$ & $0.84 \pm 0.0083$ & $0.56 \pm 0.0103$ & $0.60 \pm 0.0104$ & $0.75 \pm 0.0144$ & $0.84 \pm 0.0185$ \\
            Random Forest & $0.80 \pm 0.0073$ & $0.67 \pm 0.0116$ & $0.80 \pm 0.0073$ & $0.73 \pm 0.0098$ & $0.60 \pm 0.0158$ & $0.16 \pm 0.0315$ \\ 
            Gradient Boosting & $0.79 \pm 0.0077$ & $0.74 \pm 0.0111$ & $0.79 \pm 0.0077$ & $0.75 \pm 0.0092$ & $0.63 \pm 0.0149$ & $0.26 \pm 0.0276$ \\ 
            Ada Boost & $0.80 \pm 0.0070$ & $0.73 \pm 0.0145$ & $0.81 \pm 0.0070$ & $0.74 \pm 0.0095$ & $0.65 \pm 0.0156$ & $0.20 \pm 0.0346$ \\ 
            XGBoost & $0.78 \pm 0.0081$ & $0.73 \pm 0.0115$ & $0.78 \pm 0.0081$ & $0.75 \pm 0.0097$ & $0.56 \pm 0.0166$ & $0.26 \pm 0.0287$ \\ \hline
            \multicolumn{7}{|c|}{\textbf{BioClinical BERT + DocBERT}} \\
            \hline
            Logistic Regression (w/o multicollinearity) & $0.62 \pm 0.0099$ & $0.77 \pm 0.0100$ & $0.62 \pm 0.0099$ & $0.66 \pm 0.0091$ & $0.67 \pm 0.0125$ & $0.64 \pm 0.0190$ \\
            Random Forest & $0.80 \pm 0.0072$ & $0.72 \pm 0.0137$ & $0.80 \pm 0.0072$ & $0.74 \pm 0.0099$ & $0.60 \pm 0.0143$ & $0.20 \pm 0.0316$ \\
            Gradient Boosting & $0.71 \pm 0.0084$ & $0.70 \pm 0.0108$ & $0.71 \pm 0.0084$ & $0.70 \pm 0.0091$ & $0.62 \pm 0.0140$ & $0.28 \pm 0.026$ \\
            AdaBoost & $0.79 \pm 0.0073$ & $0.74 \pm 0.0108$ & $0.79 \pm 0.0073$ & $0.75 \pm 0.0094$ & $0.65 \pm 0.0157$ & $0.28 \pm 0.0291$ \\
            XGBoost & $0.79 \pm 0.0073$ & $0.74 \pm 0.0101$ & $0.79 \pm 0.0073$ & $0.76 \pm 0.0090$ & $0.64 \pm 0.0143$ & $0.29 \pm 0.0281$ \\ \hline
            \multicolumn{7}{|c|}{\textbf{BioBERT + DocBERT}} \\
            \hline
            Logistic Regression & $0.54 \pm 0.0108$ & $0.78 \pm 0.0107$ & $0.54 \pm 0.0108$ & $0.58 \pm 0.0101$ & $0.62 \pm 0.0157$ & $0.70 \pm 0.02$ \\
            Random Forest & $0.81 \pm 0.0072$ & $0.67 \pm 0.0116$ & $0.81 \pm 0.0072$ & $0.73 \pm 0.0099$ & $0.60 \pm 0.0147$ & $0.18 \pm 0.007$ \\
            Gradient Boosting & $0.79 \pm 0.0079$ & $0.75 \pm 0.0111$ & $0.79 \pm 0.0079$ & $0.76 \pm 0.0096$ & $0.68 \pm 0.0139$ & $0.31 \pm 0.0172$ \\ 
            Ada Boost & $0.78 \pm 0.0080$ & $0.73 \pm 0.0112$ & $0.78 \pm 0.0080$ & $0.74 \pm 0.0096$ & $0.65 \pm 0.0151$ & $0.31 \pm 0.0134$ \\ 
            XGBoost & $0.79 \pm 0.0083$ & $0.76 \pm 0.0107$ & $0.79 \pm 0.0083$ & $0.77 \pm 0.0096$ & $0.61 \pm 0.0157$ & $0.33 \pm 0.0277$ \\ \hline
            \multicolumn{7}{|c|}{\textbf{Pre-trained GatorTron Base Weights}} \\
            \hline
            Logistic Regression & $0.46 \pm 0.2944$ & $0.36 \pm 0.3186$ & $0.46 \pm 0.2944$ & $0.36 \pm 0.3165$ & $0.68 \pm 0.0234$ & $0.52 \pm 0.294$ \\
            Random Forest & $0.80 \pm 0.0161$ & $0.73 \pm 0.0789$ & $0.805 \pm 0.0152$ & $0.74 \pm 0.0266$ & $0.66 \pm 0.0520$ & $0.25 \pm 0.0604$ \\
            Gradient Boosting & $0.77 \pm 0.0275$ & $0.75 \pm 0.035$ & $0.77 \pm 0.0275$ & $0.76 \pm 0.0296$ & $0.68 \pm 0.0389$ & $0.34 \pm 0.089$ \\ 
            Ada Boost & $0.70 \pm 0.0362$ & $0.75 \pm 0.0313$ & $0.70 \pm 0.0362$ & $0.71 \pm 0.0291$ & $0.64 \pm 0.0456$ & $0.49 \pm 0.1120$ \\
            XGBoost & $0.77 \pm 0.0284$ & $0.73 \pm 0.0462$ & $0.77 \pm 0.0325$ & $0.74 \pm 0.0313$ & $0.62 \pm 0.0457$ & $0.33 \pm 0.0837$ \\ \hline
            \multicolumn{7}{|c|}{\textbf{Pre-trained Llama 8B Weights}} \\
            \hline
            Logistic Regression & $0.55 \pm 0.2733$ & $0.49 \pm 0.2838$ & $0.55 \pm 0.2733$ & $0.47 \pm 0.2991$ & $0.51 \pm 0.0319$ & $0.44 \pm 0.2751$ \\ 
            Random Forest & $0.80 \pm 0.0158$ & $0.72 \pm 0.0831$ & $0.80 \pm 0.0158$ & $0.72 \pm 0.0185$ & $0.59 \pm 0.0576$ & $0.22 \pm 0.0424$ \\
            Gradient Boosting & $0.74 \pm 0.0565$ & $0.71 \pm 0.0464$ & $0.74 \pm 0.0565$ & $0.71 \pm 0.0363$ & $0.60 \pm 0.0521$ & $0.31 \pm 0.0997$ \\
            Ada Boost & $0.66 \pm 0.0967$ & $0.73 \pm 0.0419$ & $0.66 \pm 0.0967$ & $0.67 \pm 0.0753$ & $0.61 \pm 0.0458$ & $0.47 \pm 0.1586$ \\
            XGBoost & $0.68 \pm 0.0904$ & $0.69 \pm 0.0348$ & $0.68 \pm 0.0904$ & $0.68 \pm 0.0630$ & $0.55 \pm 0.0552$ & $0.36 \pm 0.1186$ \\  \hline
            \multicolumn{7}{|c|}{\textbf{Pre-trained Mistral 7B Weights}} \\
            \hline
            Logistic Regression & $0.60 \pm 0.2067$ & $0.69 \pm 0.1588$ & $0.60 \pm 0.2067$ & $0.58 \pm 0.2163$ & $0.65 \pm 0.0254$ & $0.49 \pm 0.2154$ \\
            Random Forest & $0.78 \pm 0.0253$ & $0.74 \pm 0.0435$ & $0.78 \pm 0.0253$ & $0.74 \pm 0.0234$ & $0.63 \pm 0.0398$ & $0.31 \pm 0.0681$ \\
            Gradient Boosting & $0.71 \pm 0.0440$ & $0.72 \pm 0.0255$ & $0.71 \pm 0.0440$ & $0.71 \pm 0.0316$ & $0.61 \pm 0.0367$ & $0.42 \pm 0.0838$ \\
            Ada Boost & $0.69 \pm 0.0780$ & $0.74 \pm 0.0256$ & $0.69 \pm 0.0781$ & $0.70 \pm 0.0624$ & $0.63 \pm 0.0350$ & $0.51 \pm 0.0946$ \\
            XGBoost & $0.69 \pm 0.0502$ & $0.72 \pm 0.0259$ & $0.69 \pm 0.0502$ & $0.70 \pm 0.0358$ & $0.60 \pm 0.0440$ & $0.44 \pm 0.0869$ \\ \hline
        \end{tabular}
    }

    \label{tab:benchmark-results}
\end{table*}

With GatorTron embeddings, Random Forest recorded the highest accuracy (0.80 ± 0.0161 ) and recall (0.805 ± 0.0152), paired with a strong F1 score of 0.74 ± 0.0266. Gradient Boosting closely followed, achieving an AUC-ROC of 0.68 ± 0.0389, the highest among GatorTron classifiers. 

Llama embeddings demonstrated comparable performance to GatorTron embeddings, showing slight improvements in some models like logistic regression while maintaining competitive results in others. Mistral embeddings had the lowest metrics compared to the other LLMs tested. However, it did have higher weighted specificity in AdaBoost 0.51 ± 0.0946) compared to other LLMs. 

The table depicts that even the large language models, yet being trained on millions of parameters, could not perform better than the small language models. They achieved comparable results on the classification task for predicting the onset of abnormal shock index, highlighting the need for these models to be trained for prediction tasks for better performance. This also suggests that merely increasing model size may not suffice to achieve superior outcomes; rather, it demonstrates the need for tailored training approaches across such tasks to enhance performance.

\begin{figure}[htbp]
\centerline{\includegraphics[width=1\linewidth]{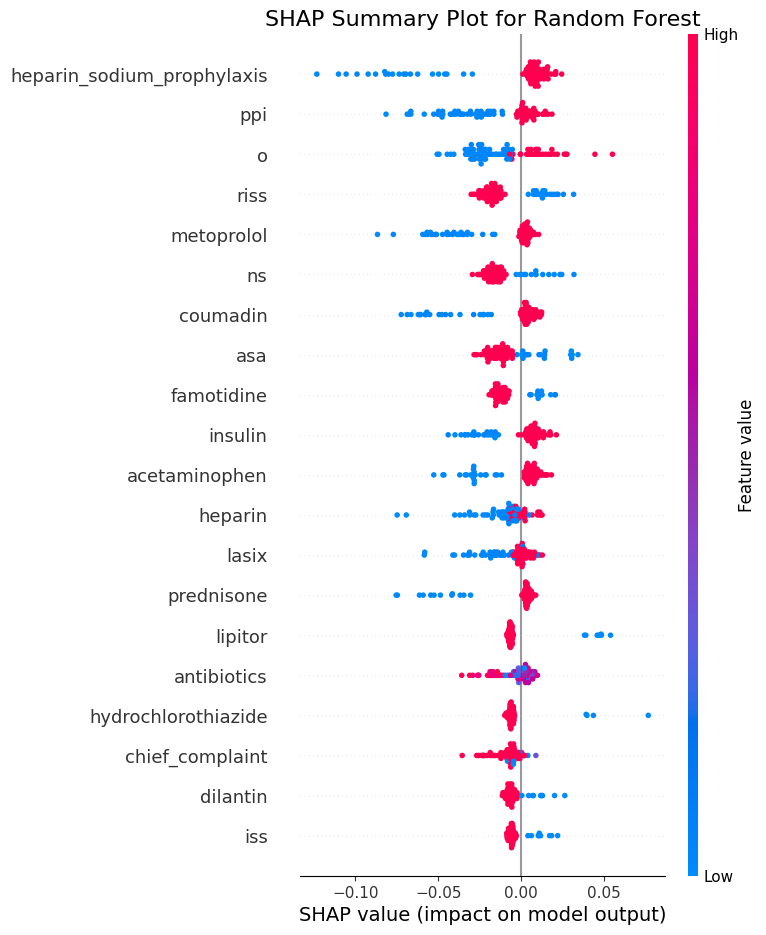}}
\caption{SHAP Summary Plot for Random Forest}
\label{rf_shap}
\end{figure}

To understand which features affected our predictions the most, we did SHapley Additive exPlanations (SHAP) \cite{shap} for GatotTron embeddings. The summary plot for our best model, Random forest based on recall has been shown in Fig.~\ref{rf_shap}.

Heparin sodium prophylaxis and coumadin strongly affect the onset of shock. Famotidine and Risperidone (riss) strongly influence the negative class, i.e., prediction of no shock. These results correlate with those in the previous work done \cite{shcokmode}, highlighting that LLMs are learning the same features as SLMs. 

\subsection{Assessing LLM Performance with and Without Fine-Tuning}

The performance of the machine learning models was evaluated under three configurations: fine-tuned with cross-entropy loss, fine-tuned with focal loss, and our pre-trained LLM, as summarized in Table 2. 

\begin{table*}[htbp]
    \caption{Performance Metrics for GatorTron Base Across Fine-Tuning Methods}
    \centering
    \resizebox{\textwidth}{!}{
        \renewcommand{\arraystretch}{1.5}
        \setlength{\tabcolsep}{5pt}
        \begin{tabular}{|c|c|c|c|c|c|c|c|}
            \hline
            \textbf{Model} & \textbf{Fine-Tuning Method} & \textbf{Accuracy} & \textbf{Precision} & \textbf{Recall} & \textbf{F1 Score} & \textbf{AUC-ROC} & \textbf{Specificity} \\
            \hline
            \multirow{3}{*}{Logistic Regression}
                & CE Loss & $0.48 \pm 0.220$ & $0.29 \pm 0.228$ & $0.48 \pm 0.220$ & $0.35 \pm 0.244$ & $0.44 \pm 0.049$ & $0.51 \pm 0.221$ \\
                & Focal Loss & $0.49 \pm 0.221$ & $0.30 \pm 0.231$ & $0.49 \pm 0.221$ & $0.36 \pm 0.246$ & $0.49 \pm 0.043$ & $0.51 \pm 0.219$ \\
                & Not Fine-Tuned & $0.46 \pm 0.2944$ & $0.36 \pm 0.3186$ & $0.46 \pm 0.2944$ & $0.36 \pm 0.3165$ & $0.68 \pm 0.0234$ & $0.52 \pm 0.294$ \\
            \hline
            \multirow{3}{*}{Random Forest}
                & CE Loss & $0.72 \pm 0.031$ & $0.64 \pm 0.122$ & $0.72 \pm 0.031$ & $0.64 \pm 0.043$ & $0.52 \pm 0.052$ & $0.32 \pm 0.056$ \\
                & Focal Loss & $0.73 \pm 0.027$ & $0.61 \pm 0.134$ & $0.73 \pm 0.027$ & $0.63 \pm 0.044$ & $0.56 \pm 0.050$ & $0.31 \pm 0.057$ \\
                & Not Fine-Tuned & $0.80 \pm 0.0161$ & $0.73 \pm 0.0789$ & $0.805 \pm 0.0152$ & $0.74 \pm 0.0266$ & $0.66 \pm 0.0520$ & $0.25 \pm 0.0604$ \\
            \hline
            \multirow{3}{*}{Gradient Boosting} 
                & CE Loss & $0.66 \pm 0.059$ & $0.62 \pm 0.082$ & $0.66 \pm 0.059$ & $0.63 \pm 0.058$ & $0.56 \pm 0.079$ & $0.36 \pm 0.083$ \\
                & Focal Loss & $0.71 \pm 0.063$ & $0.69 \pm 0.097$ & $0.71 \pm 0.063$ & $0.67 \pm 0.060$ & $0.60 \pm 0.072$ & $0.40 \pm 0.084$ \\
                & Not Fine-Tuned & $0.77 \pm 0.0275$ & $0.75 \pm 0.035$ & $0.77 \pm 0.0275$ & $0.76 \pm 0.0296$ & $0.68 \pm 0.0389$ & $0.34 \pm 0.089$ \\ 
            \hline
            \multirow{3}{*}{AdaBoost}
                & CE Loss & $0.65 \pm 0.084$ & $0.65 \pm 0.081$ & $0.65 \pm 0.084$ & $0.64 \pm 0.069$ & $0.60 \pm 0.081$ & $0.44 \pm 0.091$ \\
                & Focal Loss & $0.67 \pm 0.073$ & $0.66 \pm 0.071$ & $0.67 \pm 0.073$ & $0.65 \pm 0.062$ & $0.58 \pm 0.082$ & $0.44 \pm 0.090$ \\
                & Not Fine-Tuned & $0.70 \pm 0.0362$ & $0.75 \pm 0.0313$ & $0.70 \pm 0.0362$ & $0.71 \pm 0.0291$ & $0.64 \pm 0.0456$ & $0.49 \pm 0.1120$ \\
            \hline
            \multirow{3}{*}{XGBoost}
                & CE Loss & $0.62 \pm 0.068$ & $0.58 \pm 0.059$ & $0.62 \pm 0.068$ & $0.59 \pm 0.056$ & $0.44 \pm 0.070$ & $0.33 \pm 0.068$ \\
                & Focal Loss & $0.65 \pm 0.066$ & $0.63 \pm 0.068$ & $0.65 \pm 0.066$ & $0.63 \pm 0.054$ & $0.49 \pm 0.059$ & $0.39 \pm 0.067$ \\
                & Not Fine-Tuned & $0.77 \pm 0.0284$ & $0.73 \pm 0.0462$ & $0.77 \pm 0.0325$ & $0.74 \pm 0.0313$ & $0.62 \pm 0.0457$ & $0.33 \pm 0.0837$ \\
            \hline
        \end{tabular}
    }
    \label{tab:performance-results}
\end{table*}

Random Forest had the best accuracy (0.72 ± 0.031 and 0.73 ± 0.0273) across the different setups and the most stability. For Gradient Boosting, the focal loss configuration  enhanced specificity (0.40 ± 0.084), but other metrics remained relatively stable, with the non-fine-tuned models achieving superior precision (0.75 ± 0.0313) and recall (0.805 ± 0.0152). This suggests that while fine-tuning can address class imbalances, it may not provide consistent overall improvements. 

The F1 scores and AUC-ROC values consistently favored the non-fine-tuned configuration, indicating its robustness across metrics. Logistic Regression exhibited the lowest performance overall, with accuracy ranging from 0.46 to 0.49 and F1 scores between 0.35 and 0.36.
 
In summary, non-fine-tuned models consistently delivered superior or comparable performance across all classifiers and metrics, reinforcing the effectiveness of pre-trained embeddings in extracting meaningful features. Fine-tuning with focal loss improved specificity for some models, such as Gradient Boosting, but overall, the gains were insufficient to outweigh the stronger generalizability of non-fine-tuned approaches. These findings emphasize the utility of pre-trained embeddings for robust and efficient predictive modeling in ICU contexts.

\subsection{Exploring Variations in Fine-Tuning Parameters and Model Performance}

\begin{table*}[htbp]
    \caption{Performance Metrics for Different Dropout Models for GatorTron Base}
    \centering
    \resizebox{\textwidth}{!}{ 
        \renewcommand{\arraystretch}{1.5}
        \setlength{\tabcolsep}{5pt}
        \begin{tabular}{|c|c|c|c|c|c|c|c|}
            \hline
            \textbf{Model} & \textbf{Dropout/Epochs} & \textbf{Accuracy} & \textbf{Precision} & \textbf{Recall} & \textbf{F1 Score} & \textbf{AUC ROC} & \textbf{Specificity} \\ \hline
            Logistic Regression & 0.7/13 & $0.54 \pm 0.2145$ & $0.35 \pm 0.2226$ & $0.54 \pm 0.2145$ & $0.41 \pm 0.2376$ & $0.43 \pm 0.0443$ & $0.45 \pm 0.2159$ \\
             & 0.999999/13 & $0.49 \pm 0.2209$ & $0.30 \pm 0.2252$ & $0.49 \pm 0.2209$ & $0.36 \pm 0.2431$ & $0.43 \pm 0.0397$ & $0.50 \pm 0.2216$ \\
             & 0.9999/20 & $0.50 \pm 0.2177$ & $0.32 \pm 0.2273$ & $0.50 \pm 0.2177$ & $0.38 \pm 0.22417$ & $0.44 \pm 0.0372$ & $0.48 \pm 0.2184$ \\ \hline
             Random Forest & 0.7/13 & $0.73 \pm 0.0253$ & $0.67 \pm 0.1235$ & $0.73 \pm 0.0253$ & $0.64 \pm 0.0373$ & $0.48 \pm 0.0527$ & $0.33 \pm 0.0471$ \\ 
             & 0.999999/13 & $0.71 \pm 0.0337$ & $0.63 \pm 0.1169$ & $0.7186 \pm 0.0337$ & $0.63 \pm 0.0409$ & $0.54 \pm 0.0588$ & $0.32 \pm 0.0508$ \\ 
             & 0.9999/20 & $0.76 \pm 0.0328$ & $0.77 \pm 0.0976$ & $0.76 \pm 0.0328$ & $0.70 \pm 0.0446$ & $0.51 \pm 0.0475$ & $0.41 \pm 0.0591$ \\ \hline
            Gradient Boosting & 0.7/13 & $0.68 \pm 0.0592$ & $0.64 \pm 0.0915$ & $0.68 \pm 0.0592$ & $0.64 \pm 0.0609$ & $0.53 \pm 0.0723$ & $0.37 \pm 0.0855$ \\ 
             & 0.999999/13 & $0.68 \pm 0.0615$ & $0.64 \pm 0.0827$ & $0.68 \pm 0.0615$ & $0.65 \pm 0.0618$ & $0.60 \pm 0.0732$ & $0.39 \pm 0.0928$ \\ 
             & 0.9999/20 & $0.69 \pm 0.0614$ & $0.67 \pm 0.0690$ & $0.69 \pm 0.0614$ & $0.67 \pm 0.0545$ & $0.56 \pm 0.0628$ & $0.44 \pm 0.0698$ \\ \hline
            Ada Boost & 0.7/13 & $0.67 \pm 0.0702$ & $0.64 \pm 0.0699$ & $0.67 \pm 0.0702$ & $0.64 \pm 0.0599$ & $0.56 \pm 0.0686$ & $0.40 \pm 0.0710$ \\ 
             & 0.999999/13 & $0.64 \pm 0.0638$ & $0.64 \pm 0.0650$ & $0.64 \pm 0.0638$ & $0.64 \pm 0.0561$ & $0.57 \pm 0.0722$ & $0.45 \pm 0.0851$ \\ 
             & 0.9999/20 & $0.63 \pm 0.0762$ & $0.63 \pm 0.0651$ & $0.63 \pm 0.0762$ & $0.63 \pm 0.0639$ & $0.54 \pm 0.0513$ & $0.43 \pm 0.0690$ \\ \hline            
            XG Boost & 0.7/13 & $0.66 \pm 0.0524$ & $0.61 \pm 0.0619$ & $0.66 \pm 0.0524$ & $0.62 \pm 0.0407$ & $0.45 \pm 0.0727$ & $0.36 \pm 0.0677$ \\
             & 0.999999/13 & $0.65 \pm 0.0549$ & $0.62 \pm 0.0605$ & $0.65 \pm 0.0549$ & $0.63 \pm 0.0519$ & $0.54 \pm 0.0776$ & $0.38 \pm 0.0851$ \\
             & 0.9999/20 & $0.69 \pm 0.0656$ & $0.68 \pm 0.0595$ & $0.69 \pm 0.0656$ & $0.68 \pm 0.0544$ & $0.51 \pm 0.0501$ & $0.46 \pm 0.0507$ \\ \hline
            
        \end{tabular}
    }
    \label{table:metrics}
\end{table*}

The results of fine-tuning using focal loss across various dropout rates (0.7, 0.9999, 0.999999) have been summarized in Table 3.

For Gradient Boosting, the accuracy ranged from 0.68 to 0.69, with minimal differences across dropout rates. Precision, recall, F1 score, AUC-ROC, and specificity also showed small variations, indicating that no single dropout configuration had a clear advantage. Similarly, for AdaBoost, the performance metrics, including accuracy, precision, recall, and specificity, remained stable across dropout rates, highlighting the model’s insensitivity to dropout changes.

Random Forest displayed slightly larger variations, with accuracy ranging from 0.73 to 0.76, but other metrics such as F1 score, recall, precision, and specificity remained relatively stable, confirming the model's robustness. XGBoost also demonstrated consistent performance, with narrow variations across accuracy, precision, and recall, suggesting minimal sensitivity to dropout configurations.

Logistic Regression, while showing lower overall performance compared to ensemble methods, maintained consistent accuracy, F1 score, and AUC-ROC across dropout rates, further confirming its stability under different configurations.

These results suggest that performance metric variations are minimal across models, indicating that the models are generally robust to changes in dropout rate. However, this robustness depends on selecting an appropriate number of epochs, ensuring convergence of the loss curves.

Our preliminary results suggest that while LLMs are capable of handling the complexity of ICU data, they did not outperform SLMs. This may be due to the lack of training on tasks specifically involving the prediction of patient trajectories. Additionally, fine-tuning did not yield significant improvements, likely because of the limited size of the fine-tuning cohort. These findings highlight the importance of training LLMs on large datasets focusing on prediction tasks rather than solely cross-sectional tasks such as named entity recognition, link prediction, and phenotyping. Further refinement is essential for LLMs to meet the demands of real-world clinical applications and achieve performance comparable to traditional methods.

\section{Discussion}
The comparative analysis of LLMs and SLMs in this study reveals that despite the advanced capabilities of LLMs in processing extensive text data, they do not always perform better than SLMs in clinical prediction tasks. LLMs, like GatorTron-Base, a large model trained on medical data, and SLMs, such as BioBERT and Doc2Vec, demonstrated comparable efficacy in predicting physiological decompensation in critically ill patients. This equivalence suggests that the decision to deploy LLMs in clinical settings should be meticulously tailored to specific task requirements rather than based on a generalized model preference. The extraction of embeddings from LLMs for use in traditional machine learning models, rather than direct classification, was strategically chosen to leverage the rich contextual understanding of LLMs while harnessing the robustness and simplicity of traditional models in handling dynamic clinical data \cite {b3}. 

Random Forest emerged as the most robust classifier across GatorTron, Llama, and Mistral embeddings, achieving the highest accuracy and F1 scores. Ensemble-based classifiers, such as Random Forest and Gradient Boost, consistently outperformed simpler models like Logistic Regression in fine-tuned and non-fine-tuned setups. This highlights the importance of model complexity and effective feature selection when working with high-dimensional ICU data. Furthermore, focal loss during fine-tuning addressed class imbalances, improving recall and F1 scores compared to cross-entropy loss. However, non-fine-tuned LLMs often performed better across metrics, suggesting that pre-trained knowledge generalizes ICU data effectively without extensive task-specific optimization. These findings align with existing literature on the robustness of pre-trained embeddings for complex predictive tasks \cite{shcokmode} .

Our objective was to evaluate the role of LLMs, including GatorTron, Llama, and Mistral, in predictive applications within healthcare and compare their performance to SLMs like BioClinicalBERT and Word2Vec-Doc2Vec. While LLMs have shown potential in other tasks \cite{b4} , our results indicate that their effectiveness in predictive modeling is often limited and does not consistently surpass that of established SLMs. Previous work with SLMs has set a high benchmark in this domain, and our findings reinforce the necessity of benchmarking LLMs to understand their role in predictive healthcare tasks.

A key limitation of this study is the small dataset size, which constrained our ability to fine-tune LLMs effectively. To address this, future research should explore alternative fine-tuning methods or focus on pre-training models, specifically predictive tasks. Training LLMs directly for these tasks may enhance their performance and applicability in healthcare.

These findings are significant because they emphasize the need for critical evaluation of LLMs, including GatorTron, Llama, and Mistral, in predictive contexts. As LLMs are increasingly utilized in healthcare for tasks such as named entity recognition and clinical summarization, benchmarking their performance in predictive tasks is essential to understanding their capabilities and limitations. Comparing LLMs to SLMs ensures that we accurately determine where these models stand and identify areas for improvement in their development.

\section{Conclusion}
In this study, we assessed the efficacy of LLMs for predictive healthcare tasks, specifically shock prediction in ICU settings, and compared their performance against SLMs. Our findings highlight that while LLMs like Llama 8B, Mistral 7B, and GatorTron Base show potential in extracting and integrating complex clinical data, they did not outperform SLMs in this context. SLMs also exhibited superior specificity and robustness.  Notably, LLMs that were pre-trained on healthcare-specific data did not show significant improvements compared to their counterparts that were not pre-trained on such data.

However, our study faced limitations, such as a relatively small dataset, which restricted the efficacy of advanced fine-tuning strategies involving various loss functions and dropout techniques.

Future work should address these challenges by fine-tuning using larger, more diverse cohorts and exploring tasks like predicting patient trajectories during the training of LLMs. By building on these insights, LLMs hold promise as transformative tools for predictive applications in clinical settings, complementing the robust performance of SLMs in targeted healthcare tasks.

\bibliographystyle{IEEEtran}
\bibliography{references}

\end{document}